\documentclass[conference, 9pt]{IEEEtran}
\IEEEoverridecommandlockouts

\usepackage{cite}
\usepackage{amsmath,amssymb,amsfonts}
\usepackage{algorithmic}
\usepackage{tabularx}
\usepackage{adjustbox}
\usepackage{multirow}
\usepackage{makecell}
\usepackage{graphicx}
\usepackage{textcomp}
\usepackage{comment}
\usepackage{xcolor}
\usepackage{cleveref}
\usepackage{braket}
\usepackage{enumitem}
\usepackage{makecell}
\usepackage{url}
\usepackage{enumitem}
\usepackage{tikz}
\def\BibTeX{{\rm B\kern-.05em{\sc i\kern-.025em b}\kern-.08em
    T\kern-.1667em\lower.7ex\hbox{E}\kern-.125emX}}

\usepackage[compact]{titlesec}

\titlespacing\section{0pt}{0.3\baselineskip}{0.2\baselineskip}
\titlespacing\subsection{0pt}{0.2\baselineskip}{0.1\baselineskip}
\titlespacing\subsubsection{0pt}{0.15\baselineskip}{0.1\baselineskip}

\usepackage{fancyhdr,lipsum}
\fancypagestyle{firstpage}{
  \fancyhf{}
  \fancyhead[C]{To appear at the 31st IEEE International Symposium on On-Line Testing and Robust System Design (IOLTS), Ischia, Italy, July 2025.}
  \fancyfoot[C]{\thepage}
}

\begin{document}

\title{Neuromorphic Computing for Embodied Intelligence in Autonomous Systems: Current Trends, Challenges, and Future Directions
\vspace{-10pt}
}

\author{\IEEEauthorblockN{Alberto Marchisio and Muhammad Shafique}
\IEEEauthorblockA{eBRAIN Lab, Division of Engineering, New York University Abu Dhabi (NYUAD), Abu Dhabi, UAE\\
alberto.marchisio@nyu.edu, muhammad.shafique@nyu.edu\\ 
}
\vspace{-25pt}
}

\maketitle
\thispagestyle{firstpage}

\IEEEpubidadjcol

\begin{abstract}
The growing need for intelligent, adaptive, and energy-efficient autonomous systems across fields such as robotics, mobile agents (e.g., UAVs), and self-driving vehicles is driving interest in neuromorphic computing. By drawing inspiration from biological neural systems, neuromorphic approaches offer promising pathways to enhance the perception, decision-making, and responsiveness of autonomous platforms. This paper surveys recent progress in neuromorphic algorithms, specialized hardware, and cross-layer optimization strategies, with a focus on their deployment in real-world autonomous scenarios. Special attention is given to event-based dynamic vision sensors and their role in enabling fast, efficient perception. The discussion highlights new methods that improve energy efficiency, robustness, adaptability, and reliability through the integration of spiking neural networks into autonomous system architectures. We integrate perspectives from machine learning, robotics, neuroscience, and neuromorphic engineering to offer a comprehensive view of the state of the field. Finally, emerging trends and open challenges are explored, particularly in the areas of real-time decision-making, continual learning, and the development of secure, resilient autonomous systems.

\end{abstract}

\begin{IEEEkeywords}
Neuromorphic Computing, Embodied Intelligence, Autonomous Systems, Spiking Neural Networks.
\end{IEEEkeywords}

\section{Introduction}

Autonomous systems, such as self-driving vehicles, drones, and wearable devices, require real-time decision-making under strict energy and computational constraints. Conventional architectures often struggle to meet these demands due to their high power consumption and latency. Neuromorphic computing, inspired by the energy efficiency of the human brain, offers a viable alternative through Spiking Neural Networks (SNNs), which process information via discrete events, enabling low-power, low-latency inference~\cite{viale2021carsnn, viale2022lanesnns}. This event-driven paradigm is particularly well-suited for handling temporally sparse data, aligning naturally with event-based sensing modalities.

Neuromorphic computing lies at the intersection of robotics, neuroscience, machine learning, and hardware design~\cite{marchisio2023embedded, putra2024embodied}. However, its application to autonomous agents still requires cohesive strategies that address data encoding, model robustness, co-design of algorithms and hardware, and real-time adaptability. This paper presents a unified workflow that integrates recent advances across neuromorphic algorithms, hardware platforms, and event-based sensors, with cross-layer optimization techniques aimed at improving energy efficiency, security, reliability, and adaptability, thus laying a foundation for embodied intelligence in autonomous systems.

The key contributions of this paper are:
\vspace{-5pt}
\begin{itemize}[leftmargin=*]
    \item A comprehensive workflow for developing secure, efficient, and reliable SNNs tailored for neuromorphic deployment in autonomous agents.
    \item Overview of design and optimization techniques of SNN architectures with emphasis on robustness, energy efficiency, and secure training through structural tuning, model compression, adversarial defenses, encryption techniques, and implementation of adaptive SNN mechanisms.
    \item Identification of current challenges and open research questions in neuromorphic systems, providing insights for future exploration.
\end{itemize}









\section{Neuromorphic Computing with SNNs}

\subsection{Spiking Neural Networks}

\begin{figure}[htbp]
    \centering
    \includegraphics[width=\linewidth]{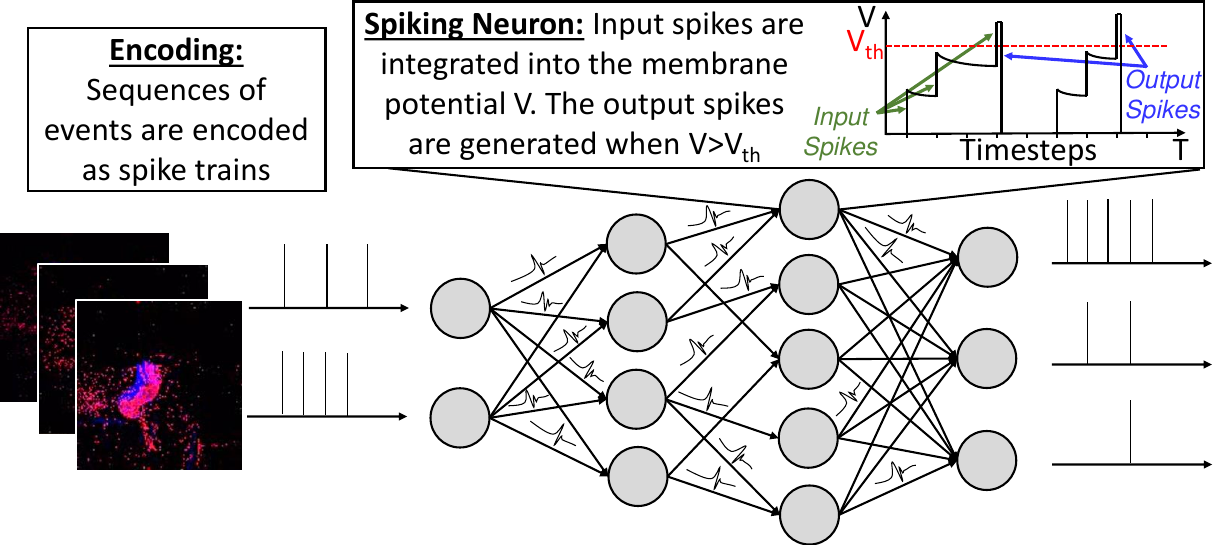}
    \caption{Basic functionality of an SNN, where input events are encoded into spike trains, and the neuron integrates these into its membrane potential.}
    \label{fig:SNN_IOLTS}
\end{figure}

Spiking Neural Networks (SNNs), often considered the third generation of neural models, emulate biological neurons more closely than traditional networks~\cite{maass1997networks}. Rather than processing continuous values, SNNs operate through discrete spikes: each neuron integrates incoming spikes into a membrane potential and fires when a threshold is exceeded (see \Cref{fig:SNN_IOLTS}). This spike-based mechanism enables efficient processing of temporal patterns, making SNNs ideal for time-sensitive tasks such as motion recognition and real-time autonomous decision-making~\cite{putra2024embodied}.

\subsection{Neuromorphic Hardware}

Neuromorphic platforms are specialized for executing SNNs efficiently by replicating structural and functional characteristics of biological systems. Examples include IBM’s TrueNorth~\cite{akopyan2015truenorth}, Intel’s Loihi~\cite{orchard2021Loihi2}, SpiNNaker, BrainScaleS~\cite{pehle2022brainscales}, and commercial chips like Brainchip Akida and Synsense DYNAP-SE~\cite{richter2024dynapse}. These systems are optimized for high computational throughput with minimal energy consumption, supporting scalable, low-power deployment of complex neural models in real-world environments.

\begin{figure*}[htbp]
    \centering
    \includegraphics[width=\textwidth]{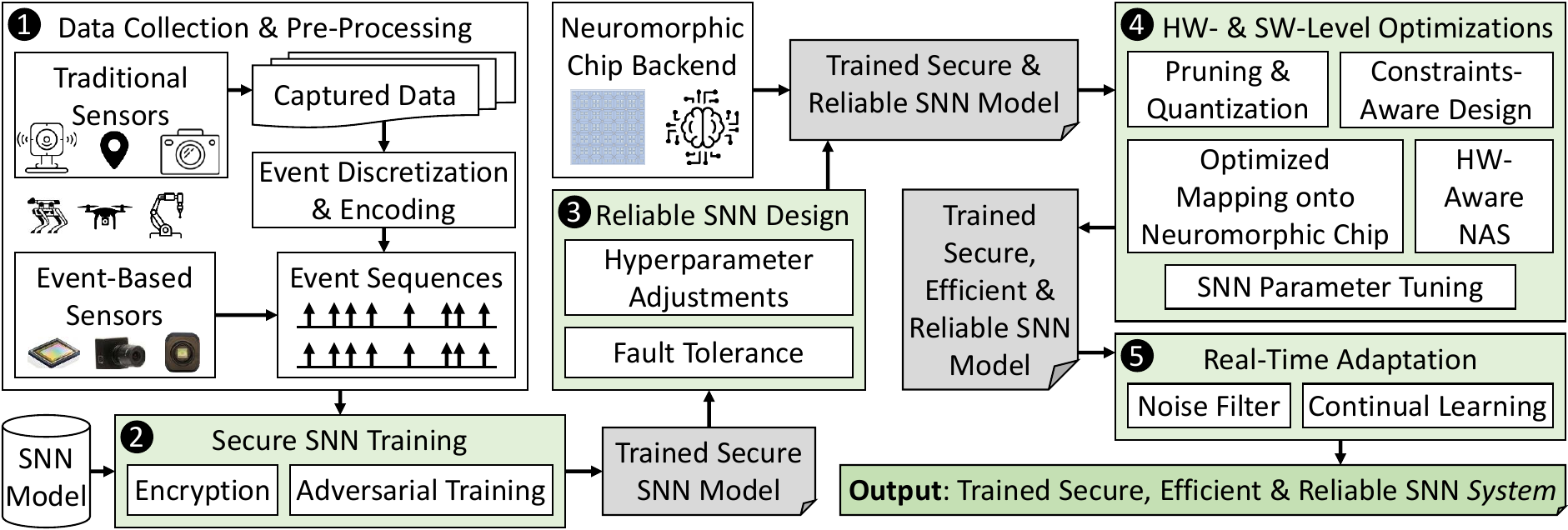}
    \caption{Overview of the proposed workflow. After collecting data from sensors and pre-processing, secure SNN training is conducted to obtain a trained secure SNN model. Afterward, reliable SNN design is conducted to obtain a reliable SNN model on the neuromorphic chip backend. Further HW- and SW-level optimizations for energy efficiency are conducted. Real-time adaptation is integrated to obtain secure, efficient, and reliable SNN systems.}
    \label{fig:Workflow}
\end{figure*}

\section{Cross-Layer Workflow for Neuromorphic-Based Embodied Intelligence in Autonomous Systems}

As shown in \Cref{fig:Workflow}, the proposed workflow begins with the acquisition of raw data through traditional or event-based sensors, followed by a pre-processing stage that includes event discretization and encoding. Yielding temporally structured event sequences suitable for spiking neural processing. Subsequently, a reliable SNN model is designed, incorporating hyperparameter tuning and fault tolerance for robustness against variability in inputs and system behavior. Afterward, a secure training pipeline combines adversarial training with encryption mechanisms to produce a resilient SNN. Next, hardware- and software-level optimizations, such as pruning, quantization, and hardware-aware NAS, are applied, guided by design constraints of the neuromorphic chip. The final step introduces real-time adaptation capabilities like continual learning and noise filtering, so the model can remain efficient under dynamic conditions. These stages result in a secure, efficient, and reliable SNN system that supports embodied intelligence in autonomous systems.

\subsection{Data Collection and Pre-Processing}

Data acquisition in neuromorphic systems often begins with event-based sensors, such as Dynamic Vision Sensors (DVS), that emit events when pixel-level brightness changes exceed a threshold~\cite{lichtsteiner2008128}. This approach enables high temporal resolution and low-latency data capture, suitable for dynamic scenes. For scenarios where traditional frame-based sensors are employed, their data can be converted into event sequences through frame differencing or threshold-based encoding, mimicking sparse asynchronous events~\cite{amir2017low}. Pre-processing these streams requires optimizing for efficient SNN representation. Techniques such as noise filtering, event denoising, and temporal smoothing help reduce redundancy and improve signal quality. Additionally, methods like time-surface generation~\cite{lagorce2016hots} and histogram-based representations~\cite{sironi2018hats} help capture spatiotemporal patterns for better SNN learning and inference.

\subsection{Secure SNN Training}

\textbf{Adversarial Training of SNNs:} Securing SNNs is critical for safe, privacy-aware deployment. A key aspect is robustness against adversarial attacks. Techniques such as adversarial training inject crafted perturbations during training to help the model learn features resilient to attacks~\cite{madry2018towards}. Recent work~\cite{ding2022snnRAT} introduces spike-aware regularization tailored to the temporal dynamics of SNNs, boosting robustness without major accuracy loss.

\textbf{Encrypted SNNs:} Encryption further protects sensitive data during inference. Homomorphic encryption schemes like the Brakerski/Fan-Vercauteren (BFV)~\cite{fan2012somewhat} enable encrypted spike processing without revealing raw data. This secures user data even in untrusted environments, though at added cost. As secure neuromorphic computing evolves, combining adversarial robustness and encryption is becoming essential for trustworthy deployment.

\begin{figure}[htbp]
    \centering
    \includegraphics[width=\linewidth]{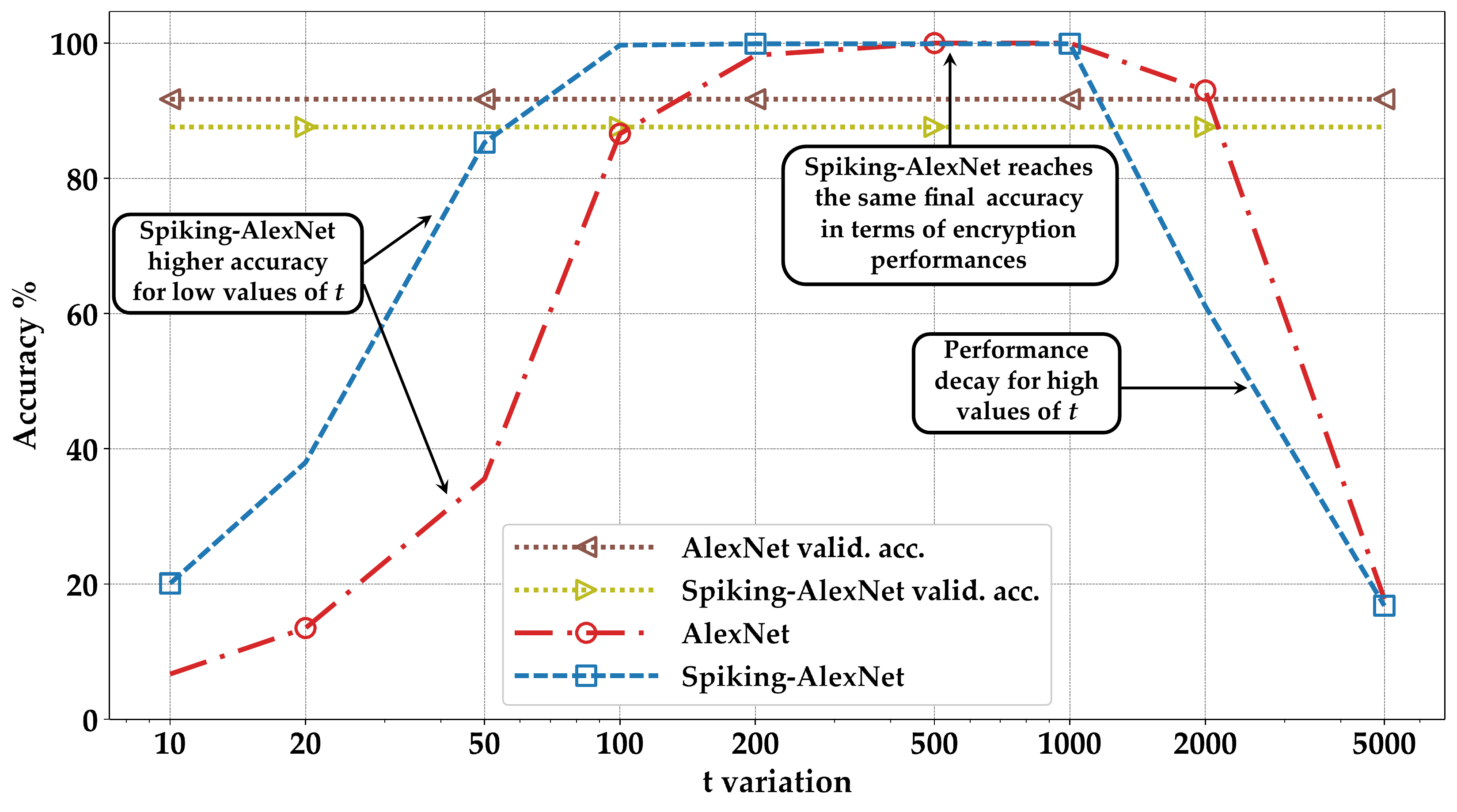}
    \caption{Accuracy comparison between AlexNet and Spiking-AlexNet for the FashionMNIST dataset, when the models are encrypted using the BFV scheme, varying the plaintext modulus $t$. Figure adapted from~\cite{nikfam2023homomorphic}.}
    \label{fig:SpyKing_accuracy_equal_fashion_alexnet}
\end{figure}

As shown in \Cref{fig:SpyKing_accuracy_equal_fashion_alexnet}, the Spiking-AlexNet maintains competitive accuracy under encryption compared to the conventional AlexNet model. It also shows improved resilience in low-modulus regimes where quantization effects dominate, suggesting better alignment of SNNs with encrypted computation constraints~\cite{nikfam2023homomorphic}.

\subsection{Reliable SNN Design}

\begin{figure}[t!]
    \centering
    \includegraphics[width=\linewidth]{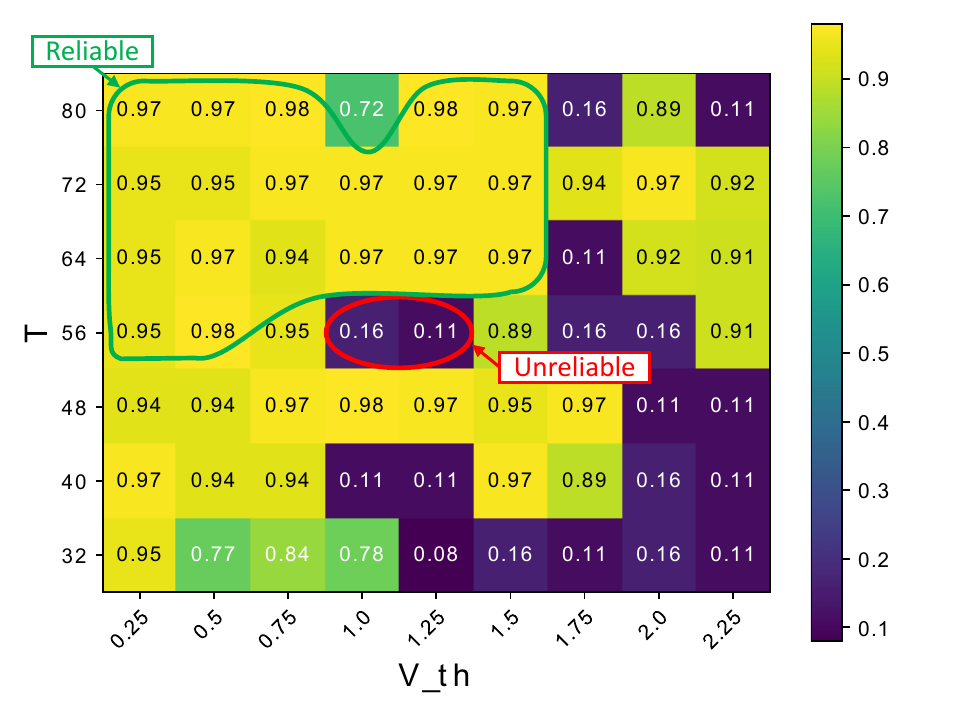}
    \vspace{-18pt}
    \caption{Robustness comparison of 4-layer SNNs on MNIST, varying the voltage threshold $V_{th}$ and time window $T$. Figure adapted from~\cite{el2021securing}.}
    \label{fig:SNN_Structural_Parameters}
\end{figure}

\textbf{SNN Hyperparameter Adjustments:} SNNs possess unique structural characteristics that are tunable for robustness. A recent study highlights how structural parameters like voltage threshold $V_{th}$ and spike integration window $T$ affect model resilience~\cite{el2021securing}. Adjusting these parameters changes spiking dynamics, affecting susceptibility to perturbations. Lower thresholds increase firing and robustness at the cost of energy, while longer time windows improve stability. \Cref{fig:SNN_Structural_Parameters} shows how proper tuning boosts robustness on MNIST with minimal architectural changes.

\begin{figure*}[htbp]
    \centering
    \includegraphics[width=\textwidth]{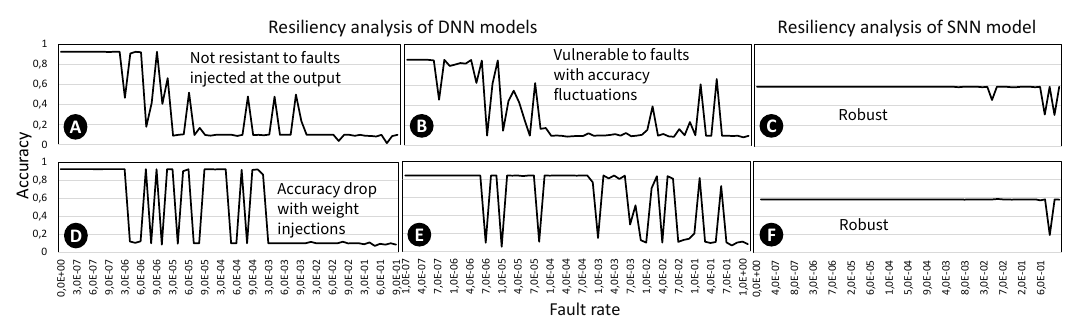}
    \caption{Comparison of test accuracy for different models under fault injection. A: Output injection on the ResNet18 for the CIFAR-10 dataset. B: Output injection on the VGG11 for the CIFAR-10 dataset. C: Output injection on a 4-layer SNN on the DVS-Gesture dataset. D: Weight injection on the ResNet18 for the CIFAR-10 dataset. E: Weight injection on the VGG11 for the CIFAR-10 dataset. F: Weight injection on a 4-layer SNN on the DVS-Gesture dataset. Figure adapted from~\cite{colucci2022enpheeph}.}
    \label{fig:enpheeph_results}
\end{figure*}

\textbf{SNN Fault Tolerance:} Beyond hyperparameters, model reliability under faults is crucial, especially on embedded neuromorphic systems. The \textit{enpheeph} framework~\cite{colucci2022enpheeph} analyzes SNN robustness via fault injection. As seen in \Cref{fig:enpheeph_results}, SNNs outperform ResNet18 and VGG11 in maintaining accuracy under both output and weight faults on CIFAR-10 and DVS-Gesture. This highlights the inherent fault tolerance of SNNs due to their sparse and distributed activation.

\subsection{HW- \& SW-Level Optimizations for SNNs}

\textbf{SNN Compression:} Compression techniques such as quantization and pruning are crucial for reducing energy and memory demands of SNNs, especially in edge devices. Reducing weight count and precision reduces energy use with minimal performance loss. The SNN4Agents framework~\cite{putra2024snn4agents} shows that a 10-bit CarSNN model preserves high accuracy on NCARS, with accuracy only dropping significantly below 8 bits, highlighting the trade-off between bitwidth and performance.

\begin{figure}[htbp]
    \centering
    \includegraphics[width=\linewidth]{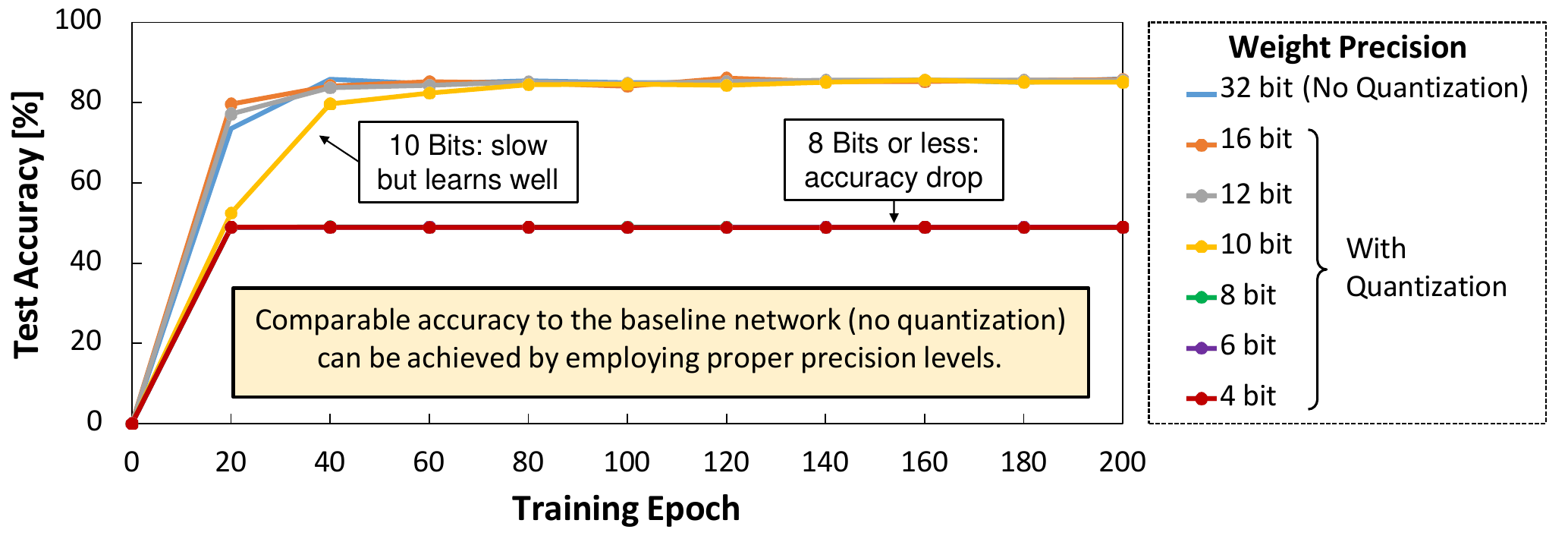}
    \caption{Accuracy of the CarSNN model on the NCARS dataset across different precision levels (from 32 bits to 4 bits). Figure adapted from~\cite{putra2024snn4agents}.}
    \label{fig:SNNforAutoAgents_QuantizeExp}
\end{figure}

\begin{figure}[htbp]
    \centering
    \includegraphics[width=\linewidth]{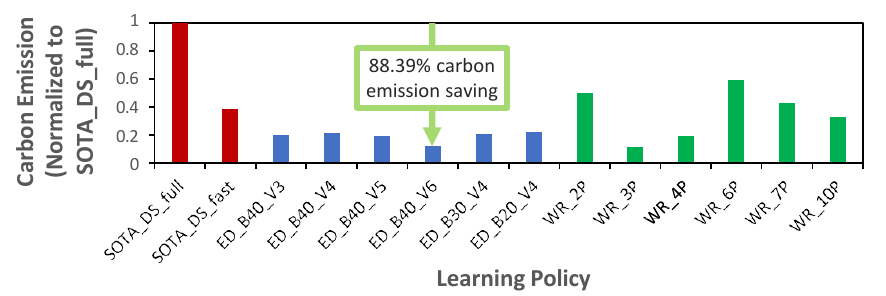}
    \caption{Exploration of different learning rate policies (exponential decay with different values of batch size and voltage threshold, warm restarts with different numbers of peaks) for the CarSNN model on the NCARS dataset and their impact on the carbon emission, compared to the baseline. Figure adapted from~\cite{bano2024fastspiker}.}
    \label{fig:FastSpiker_Results_Carbon}
\end{figure}

\textbf{Optimized SNN Learning Rate Policies:} Training efficiency is key for sustainable SNN development. FastSpiker~\cite{bano2024fastspiker} leverages learning rate strategies such as exponential decay and cosine warm restarts for faster training without sacrificing accuracy. As shown in \Cref{fig:FastSpiker_Results_Carbon}, these methods offer strong gains in efficiency, notably reducing carbon emissions.

\begin{figure}[htbp]
    \centering
    \includegraphics[width=\linewidth]{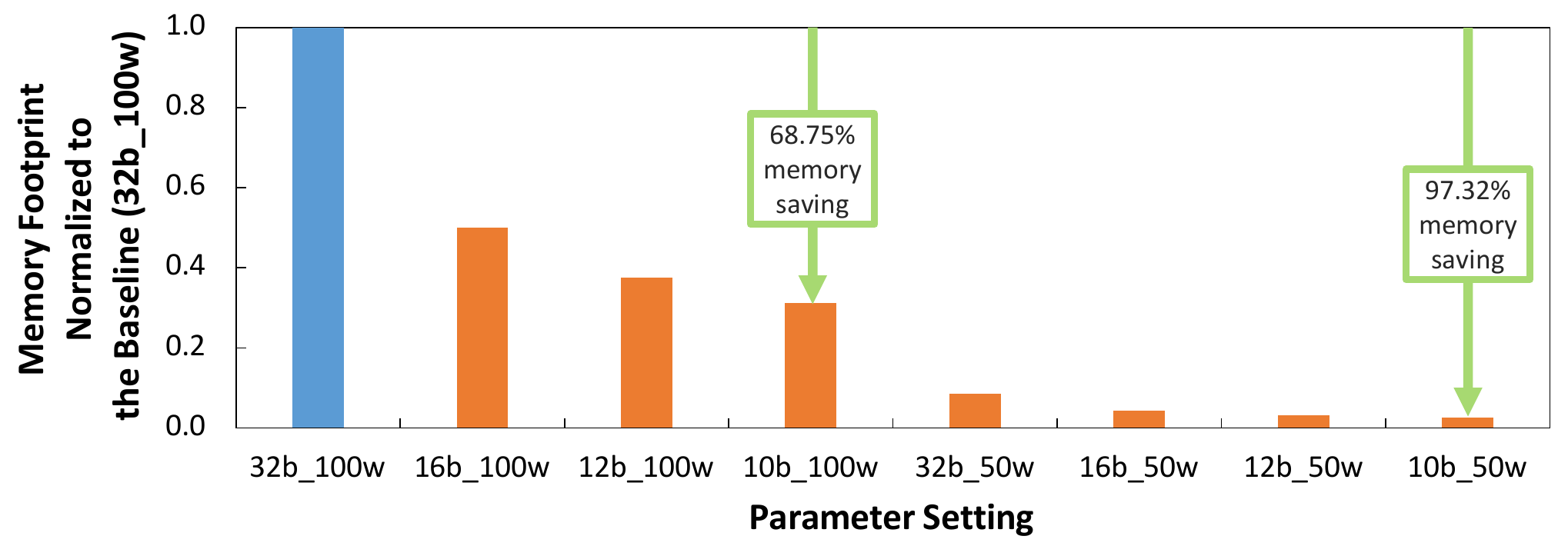}
    \caption{Exploration of different parameter configurations (number of bits and size of the attention window) of the CarSNN model on the NCARS dataset and their impact on the SNN memory footprint, compared to the baseline. Figure adapted from~\cite{putra2024snn4agents}.}
    \label{fig:SNNforAutoAgents_Result_Memory}
\end{figure}

\textbf{HW-Aware NAS and Optimized Parameter Tuning:} Optimizing SNNs through hardware-aware NAS allows for performance-driven architecture search under memory and energy constraints~\cite{bano2024methodology}. In~\cite{putra2024snn4agents}, varying precision and attention window size in CarSNN yields up to 97.32\% memory savings (see \Cref{fig:SNNforAutoAgents_Result_Memory}). Even 16-bit configurations achieve significant compression without much performance loss, enabling scalable SNN deployment in constrained systems.

\subsection{Real-Time Adaptation of SNNs}

\textbf{SNN Noise Filters:} In dynamic environments, event-based sensors frequently produce noisy spikes that degrade SNN performance. Filtering techniques like background activity filters help suppress irrelevant input spikes. R-SNN~\cite{marchisio2021RSNN} integrates such filters to improve robustness in noisy or adversarial settings by attenuating noise before it reaches spike integration.

\begin{figure}[htbp]
    \centering
    \includegraphics[width=\linewidth]{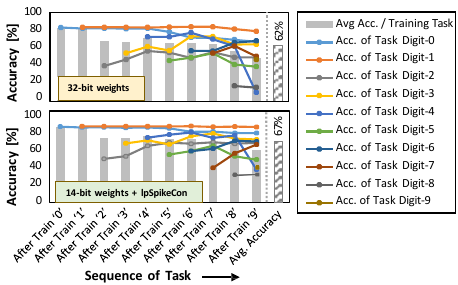}
    \caption{Accuracy of a 200-neuron SNN for the MNIST dataset, trained on different tasks, while also reducing the bitwidth. Figure adapted from~\cite{putra2022lpspikecon}.}
    \label{fig:IpSpikeCon_Results}
\end{figure}

\textbf{Continual Learning for SNNs:} Continual learning enables a model to incorporate new knowledge without forgetting old tasks. lpSpikeCon~\cite{putra2022lpspikecon} supports this via low-precision synaptic weights and online updates, enabling multi-task learning with reduced memory needs. As shown in \Cref{fig:IpSpikeCon_Results}, a 200-neuron SNN trained with lpSpikeCon retains significantly higher average accuracy, especially early ones, even under 14-bit quantization.

\section{Challenges and Open Research Questions of Neuromorphic Systems}

Despite substantial progress in deploying SNNs on neuromorphic hardware, several key challenges and open research questions persist. Tackling these issues is vital for broader adoption and effective real-world deployment.

\begin{itemize}[leftmargin=*]
    \item \textbf{Hardware Implementation and Integration:} Although SNNs enable energy-efficient processing, deploying them on neuromorphic hardware is constrained by limited precision and scalability. Bridging the algorithm-hardware gap demands co-design strategies that align model architecture with hardware capabilities~\cite{huynh2022implementing}.
    
    \item \textbf{Security and Robustness Against Backdoor Attacks:} SNNs remain vulnerable to backdoor attacks, where adversarial patterns are introduced during training. Ensuring integrity through detection and defense mechanisms remains an active area for secure deployment~\cite{jin2024data}.
    
    \item \textbf{Standardization and Benchmarking:} The absence of consistent benchmarks and evaluation protocols impedes fair comparison across neuromorphic platforms. Community-driven standard frameworks are needed to support reproducibility and accelerate progress~\cite{muir2025road}.
    
    \item \textbf{Scalability of Neuromorphic Hardware:} Current neuromorphic systems face limitations in scaling to support complex, real-world workloads. Addressing challenges related to communication overhead, power efficiency, and thermal constraints is crucial for their widespread use~\cite{kudithipudi2025neuromorphic}.
\end{itemize}

\section{Conclusion and Road Ahead}

This study outlines an end-to-end methodology for constructing secure, high-performance, and dependable SNNs optimized for neuromorphic applications in autonomous systems. The process begins with the acquisition and preprocessing of sensory data, including both traditional and event-based modalities. Subsequently, the development of robust SNN architectures is addressed through hyperparameter tuning and fault tolerance. To fortify these networks against potential threats, the integration of security measures, including adversarial robustness techniques and encryption-based safeguards, is explored. To meet stringent energy and memory limitations, the workflow incorporates model compression techniques, adaptive learning rate schedules, and architectural exploration. Real-time adaptability is addressed through noise filtering and continual learning mechanisms to ensure robustness in changing environments. Looking ahead, the advancement of neuromorphic computing for embodied intelligence will depend on scalable co-design of SNN models and hardware, efficient deployment toolchains, and standardized benchmarking. As the domain progresses, the fusion of bi-inspired neural architectures with hardware acceleration is poised to unlock a new generation of intelligent, adaptive, and energy-aware autonomous systems.


\section*{Acknowledgment}

This work was supported in parts by the NYUAD Center for Cyber Security (CCS), funded by Tamkeen under the NYUAD Research Institute Award G1104, and the NYUAD Center for Artificial Intelligence and Robotics (CAIR), funded by Tamkeen under the NYUAD Research Institute Award CG010.

\bibliographystyle{ieeetr}
\bibliography{main}

\end{document}